\documentclass[conference]{IEEEtran}
\IEEEoverridecommandlockouts
\usepackage{cite}
\usepackage{amsmath,amssymb,amsfonts}
\usepackage{algorithmic}
\usepackage{graphicx}
\usepackage{textcomp}
\usepackage{xcolor}
\usepackage{float}
\usepackage{comment}
\usepackage{tabularx}
\usepackage{array}
\usepackage{dsfont}
\usepackage{tikz}
\usetikzlibrary{shapes.geometric, arrows.meta, positioning, calc,decorations.pathreplacing}

\def\BibTeX{{\rm B\kern-.05em{\sc i\kern-.025em b}\kern-.08em
    T\kern-.1667em\lower.7ex\hbox{E}\kern-.125emX}}

\begin{document}

\title{Bridging the Data Gap: Digital Twin as a New Paradigm for AI-based Radio Sensing \\

\thanks{This work has been supported by the SNS JU project 6G-DISAC under the EU’s Horizon Europe research and innovation program under Grant Agreement No 101139130. Accepted at EUSIPCO 2026.}}

\author{
    \IEEEauthorblockN{
    Éloi Sainte-Beuve{\textsuperscript{*,$\dagger$}}, 
    Guillaume Larue{\textsuperscript{*}},
    Louis-Adrien Dufrène{\textsuperscript{*}}, 
    Quentin Lampin{\textsuperscript{*}},
    Ali Al Khansa{\textsuperscript{*}}},
    \IEEEauthorblockA{
    *Orange Research, France\\
    $\dagger$IMT Atlantique, France (former affiliation)\\
    Email: \{eloi.sainte-beuve, guillaume.larue, louisadrien.dufrene, quentin.lampin, ali.alkhansa\}@orange.com}
}

\maketitle

\begin{abstract}
We present a methodology that places a 3D digital twin (DT) of the environment as the main enabler behind the development of radio sensing at scale. The DT acts as a world model, providing geometry, materials, and transmitter/receiver placements to a ray-tracing engine that generates time-indexed channel impulse responses (CIRs) for large numbers of plausible scenes (moving people and objects, layout variants, seasonal/weather conditions, etc). From these synthetic sequences, we train a sequential neural network that maps CIR time series to spatial occupancy estimates, enabling device-free localization (DFL) without instrumented targets. We posit that sensing is best approached as an \emph{environment-conditioned} learning problem: rather than seeking a single global model, we advocate training or fine-tuning local models specialized to a site-specific DT. As a first experiment, we introduce a novel State Space Model architecture, trained and evaluated across multiple room geometries. The localization performances obtained demonstrate the potential of the approach.
\end{abstract}


\section{Introduction}

Telecommunication networks continuously illuminate the environment with radio waves. Beyond their primary role of carrying information, these waves bear rich signatures of geometry, materials, motion, and occupancy imprinted by their propagation through the environment. Integrated sensing and communication (ISAC) exploits this duality to enable sensing tasks, such as device-free localization (DFL), human activity monitoring, and scene interpretation, without deploying dedicated radar hardware or requiring targets to carry active devices \cite{6GDISAC}. For network operators with already-deployed sub-6~GHz infrastructure, ISAC offers the prospect of new services at marginal cost \cite{ZCC26}.

In modern systems (Wi‑Fi, 5G), channel estimates are commonly available in the form of channel state information (CSI) or channel impulse responses (CIR). While traditional statistical channel models used in simulation describe average or aggregate propagation behavior, the realized channel in a given deployment is a fingerprint of its environment: multipath components arising from reflection, scattering, and diffraction depend on object placement, materials, and motion. Therefore, leveraging CSI/CIR for radio sensing to extract local environmental information is a natural and broadly applicable strategy.

Early device-free systems used received signal strength indicator (RSSI) to detect presence and track intruders in Wi‑Fi environments, demonstrating feasibility but suffering from low spatial resolution and sensitivity to environmental changes \cite{YMA07}. Subsequent work shifted to finer channel descriptors CSI/CIR to capture per‑subcarrier amplitude/phase, enabling subtle motions detection, such as respiration, and through‑wall sensing \cite{WZM16,AK13}. More recently, radio frequency (RF) sensing has been formulated as a data-driven learning problem on CSI/CIR, with deep models, autoencoders, and attention-augmented hybrids, achieving strong performance in activity/gesture recognition and fingerprinting/localization, with surveys reporting improved robustness and cross-environment transfer \cite{WGM17,HZW23,NKH21,NLL25}. However, most of these approaches face practical limitations preventing scalable deployment:
\begin{itemize}
    \item \textbf{Analytical models} (e.g., Fresnel zones, line-of-sight/free-space assumptions) overly simplify actual multipath propagation and degrade in cluttered, dynamic environments, necessitating data-driven approaches.
    \item \textbf{Data-driven models} require training data that varies significantly across environments due to differences in local geometry, materials, and scenarios. CSI/CIR statistics are strongly conditioned on these factors, making broad out-of-domain generalization unlikely and rendering a single global model that works ``everywhere" elusive, thus demanding site-specific training or fine-tuning. 
    \item \textbf{Data collection} through site-specific measurement campaigns, as in fingerprinting approaches, becomes prohibitively labor-intensive when scaling across multiple sites, and require repeated data collection and retraining as layouts and usage patterns evolve.
\end{itemize}


Consequently, we advocate for a digital-twin (DT) in-the-loop methodology: for each site of interest, instantiate a 3D DT with the known static geometry and associated materials; render physics-based radio propagation to synthesize CIR sequences under diverse, plausible scenarios (occupancy, trajectories, furniture layouts, seasonal factors, etc.); and train sequential models that map CIR time series to spatial occupancy. The DT yields de facto labeled data at scale and preserves the site-specific multipath structure. One approach would be to train a coarse general model across multiple DTs, then fine-tune it locally for each specific site. 


Related DT‑in‑the‑loop sensing approaches place the RF simulator itself inside a differentiable inference loop, using inverse or differentiable ray tracing to backpropagate through the DT and directly optimize scene or pose parameters from CSI/CIR. For example, InverTwin addresses path‑space discontinuities and RF phase periodicity via tailored differentiation and surrogate models, enabling stable gradient‑based updates of DT parameters, while RayLoc performs fully differentiable wireless ray tracing for indoor localization with loss smoothing to mitigate sparse or unstable gradients \cite{XJK25,XTT25}. In spirit, these methods and ours share the idea of keeping a site‑specific DT in the loop; however, our approach is learning‑first: we use the DT primarily as a generator of labeled, site‑specific CIR sequences and train DT‑conditioned sequential models that exploit temporal consistency and environmental context to infer probabilistic occupancy and positions, rather than optimizing the DT or performing gradient‑based inverse rendering at test time.


The remainder of the paper details this methodology and presents a partial proof-of-concept (PoC) application, with emphasis on synthetic data generation and model training, while deferring real-world deployment evaluation to future work. In Section~\ref{section:framework}, we present the complete DT-in-the-loop methodology for site-specific sensing. In Section~\ref{section:system_model}, we propose our PoC where we detail the DT-based synthetic data generation pipeline, as well as model architecture and training aspects. In Section~\ref{section:results}, we evaluate the trained models on the proposed scenarios. Section~\ref{section:conclusion} discusses limitations and future deployment considerations.

\section{General Digital-Twin-in-the-Loop Framework}\label{section:framework}
We approach radio sensing as an environment-conditioned sequence learning problem: a site-specific DT, containing the 3D geometry and known RF infrastructure (e.g., access points (APs), localized user equipments (UEs) and their antenna poses/patterns), is paired with a radio sensing model based on machine learning. During training, the model consumes synthetic, time-indexed CIRs rendered from the DT; during deployment, it consumes measured CIRs associated with the current DT state. The model maps CIR sequences to probabilistic occupancy/position fields, thereby updating the DT’s dynamic state. This leverages structure beyond fundamental resolution limits: although frequency and bandwidth bound theoretical resolution, targets move continuously (no “teleportation”) and follow place-specific constraints (e.g., pedestrians on sidewalks, vehicles on roads). Processing environment-conditioned, continuous-time sequences enables the model to exploit the DT site-specific context and spatial constraints.



This framework is now practically feasible due to three key technological enablers: \textbf{(i)} mature 3D tooling and content sources for site DTs (3D mobile LIDAR capture, OpenStreetMap, Building Information Modeling, etc.), \textbf{(ii)} GPU-accelerated ray-tracing frameworks for large-scale CIR synthesis, and \textbf{(iii)} commodity hardware/software for training deep sequence models. Together, these make it practical to author site DTs, generate labeled datasets at scale, and deploy per-site models.

\subsection{Training: Synthetic Data Generation using a Digital Twin} 
During training, the DT is initialized with static geometry and RF infrastructure. Plausible dynamics (e.g., trajectories of people and vehicles, door schedules, layout variants, seasonal foliage) are sampled, and the corresponding CIR streams are rendered by an RF simulator (e.g., ray tracing). A sequence model is trained to map CIR time series to probabilistic outputs (e.g., 2D/3D occupancy heatmaps). Ground-truth occupancy/positions are read directly from the DT state, avoiding manual annotation. The same framework can be extended to predict additional attributes (e.g., material or object type). Because learning is conditioned on a site’s channel and occupancy distributions, broad zero-shot transfer to new environments is not expected, and few-shot fine-tuning is likely needed. This approach can be extended to 3D scene modeling (voxel occupancy) to feed and refine the DT \cite{RAG24}.

\subsection{Deployment: Continuous Sensing Loop}
Per-site models run with their DTs for continuous monitoring. The DT maintains both static and dynamic occupancies: tracks are updated in real time from sensing, while the DT also supplies priors for sensing. Dedicated anchor nodes (UEs with known pose in the DT) can be deployed to improve sensing performance in specific locations and can also be used during training. As targets approach DT borders, predicted positions hand over to neighboring tiles, and spatial priors help maintain consistency at scale \cite{WZN25}. Crucially, we view the DT not as a one-time offline training resource but as a continuously updated world model: real-world sensing refines the DT, which in turn improves future predictions.

\section{Proof-of-Concept Implementation}\label{section:system_model}

Building on the methodology outlined above, this section details our PoC implementation: we describe the considered DT environments, formalize the CIR‑based occupancy inference problem, and present the dataset construction, model design, and evaluation metrics. Evaluation on real-world deployment CIR data will be addressed in follow-up studies.


\begin{figure*}[htbp]
    \centering
    \begin{minipage}[t]{0.48\textwidth}
        \centering
        \includegraphics[scale=0.27]{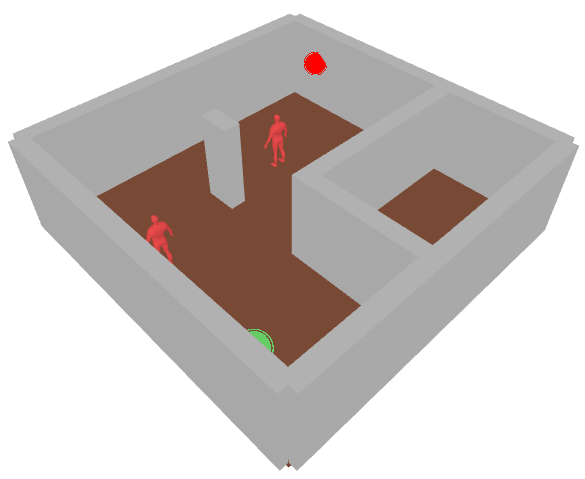}
    \end{minipage}
    \hfill
    \begin{minipage}[t]{0.48\textwidth}
        \centering
        \includegraphics[scale=0.295]{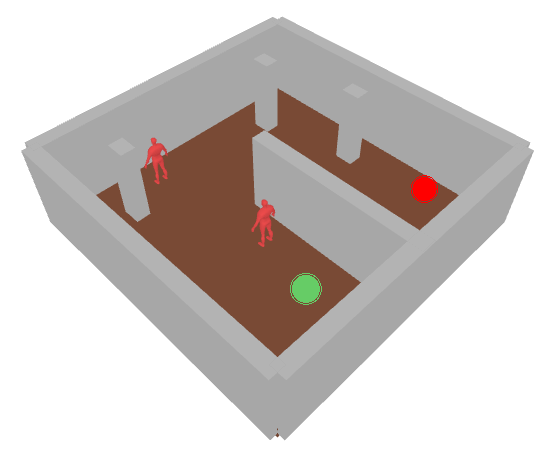}
    \end{minipage}
    \caption{\textbf{Test environments:} \textbf{(left)} asymmetric square room with blind zone, \textbf{(right)} U‑shaped room with added geometric complexity and non line of sight.}
    \label{fig:test_rooms}
\end{figure*}

\subsection{Digital Twin Environments}

Two synthetic environments, presented in Figure \ref{fig:test_rooms}, are considered. They are purpose‑built test rooms rather than reconstructions of specific real spaces. The rooms are described as 3D meshes with coarse material tags (e.g., concrete, wood).

Per room, we define a fixed couple of transmitter and receiver. During training, the receiver observes sequential complex baseband CIR of length \(K\) taps, producing a sequence of length $T$ CIRs, \(\mathcal{H}_{1:T} = [h_1,\dots,h_T]\) with \(h_t \in \mathbb{C}^{K}\). To isolate the learning problem, we adopt idealized measurements: no additive noise is injected, so the CIRs are the direct outputs of the Sionna Ray‑Tracing (Sionna RT) engine \cite{HAC23}.

In the current study, the task is to localize characters moving in the room by mapping motion‑induced multipath perturbations to occupancy. From each CIR time sequence, a model \(I_\varphi\) predicts a probabilistic occupancy heatmap on a uniform \(H\times W\) discrete floor grid at time \(t\), \(\tilde{y}_t \in [0,1]^{H\times W}\), allowing multiple simultaneous occupants. Ground‑truth maps are rasterized from the DT state at time \(t\), yielding the binary ground-truth heatmap \(b_t \in \left\{0,1\right\}^{H \times W}\). Formally,
\begin{equation}
    I_\varphi : \mathbb{C}^{K \times T} \rightarrow [0,1]^{H \times W \times T}, 
    \quad \tilde{\mathbf{y}} = I_\varphi\bigl(\mathcal{H}_{1:T}\bigr),
\end{equation}
where the parameters \(\varphi\) are learned by gradient descent to minimize a loss between \(\tilde{y}_t\) and \(b_t\).

\subsection{Dataset and Preprocessing}

We generate sequences of complex baseband CIRs together with synchronized 2D occupancy maps. The left panel of Figure \ref{fig:rssi_blur} shows an example ground-truth heatmap for the square room. Characters' starting locations are random in the room, and they move along straight paths at constant speed. Their positions labels are subsequently rasterized.

To stabilize learning and reduce spurious correlations, we apply two preprocessing steps.

\subsubsection{Baseline Removal}
For each time step $t$ and tap $k$, we subtract the empty-room CIR to suppress static multipath tied to room geometry. This subtraction substantially decreases spatial correlation between CIRs acquired with different occupant positions. Similarly, using sequences rather than single snapshots helps resolve position ambiguities that arise when instantaneous CIRs are highly correlated across different configurations.

\subsubsection{RSSI-based Label Blurring}\label{subsubsection:label_blurring}
\begin{figure}[htbp]
    \centering
    \begin{tikzpicture}
        \node[anchor=south west, inner sep=0] (image) at (0,0) {
            \includegraphics[width=\linewidth]{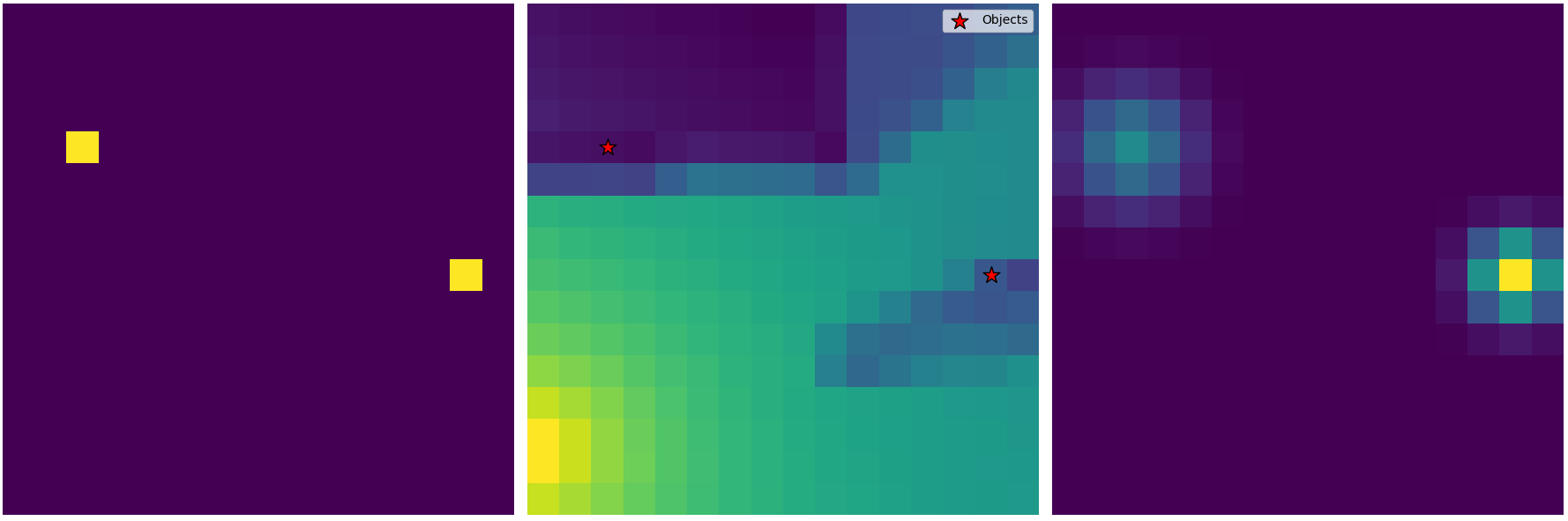}
        };
        
        \node[anchor=north, 
          fill=white,           
          draw=black,           
          rounded corners=3pt,  
          inner sep=3pt,        
          font=\scriptsize,    
          minimum width=1.5cm
         ] at (5.1, 2.92) {Targets};
         \draw[red, thick, fill=white] (4.55, 2.72) circle (3pt);
         \draw[red, thick, fill=white] (3.45, 2.12) circle (3pt);
         \draw[red, thick, fill=white] (5.6, 1.4) circle (3pt);
     
    \end{tikzpicture}
    \caption{Square room. \textbf{Left:} binary ground-truth occupancy. \textbf{Center:} RSSI map from Sionna RT. \textbf{Right:} RSSI-adaptive blurred label used for training. Strong-signal regions produce sharper targets; weak-signal regions are softened.}
    \label{fig:rssi_blur}
\end{figure}
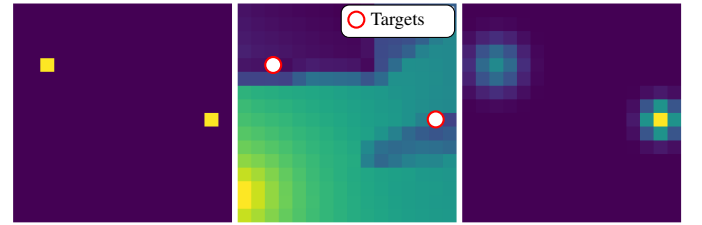

Motivated by the intuition that the model should be less penalized where the object is intrinsically harder to detect (e.g., in low-RSSI regions), we implement RSSI-adaptive label blurring. Specifically, we soften targets with a Gaussian kernel whose width decreases with the received power at the true location (computed as empty-room RSSI), reflecting higher localization confidence where the link is stronger:
\begin{equation}
    y_t \;=\; b_t * G_{\sigma}, 
    \quad G_{\sigma}[i,j] \;=\; \frac{1}{2\pi\sigma^2}\exp\!\left(-\frac{i^2+j^2}{2\sigma^2}\right),
\end{equation}
where $*$ denotes 2D convolution. So high RSSI yields a narrow spread and low RSSI a wider one. Finally, we renormalize to keep targets in $[0,1]$. The empty square room RSSI and a resulting blurred label are shown in Figure~\ref{fig:rssi_blur}. The blurred heatmaps are normalized such that \(\sum_{i,j}y_t[i,j]=1\).


\subsection{Proposed Model: Structured State-Space Approach}

State-space models (SSMs) capture the evolution of a latent state that summarizes past information and drives future predictions. In their classical form, they couple a state update with an observation map. Modern neural SSMs parameterize these operators to learn long-range temporal dependencies efficiently while remaining streamable and memory-frugal. This bridges control-theoretic formulations with deep sequence modeling, enabling stable and scalable processing of long sequences and offering a perspective closely related to, yet often more efficient than, transformers \cite{GGR22}.

\begin{figure}[t]
    \centering
    \begin{tikzpicture}[
        node distance=0.6cm,
        input_box/.style={rectangle, rounded corners=4pt, minimum width=1cm, minimum height=0.5cm, align=center, draw=black, fill=green!20,font=\scriptsize},
        process_box/.style={rectangle, rounded corners=4pt, minimum width=1cm, minimum height=0.5cm, align=center, draw=black, fill=blue!20,font=\scriptsize},
        core_box/.style={rectangle, rounded corners=4pt, minimum width=1cm, minimum height=0.5cm, align=center, draw=black, fill=orange!25,font=\scriptsize},
        decoder_box/.style={rectangle, rounded corners=4pt, minimum width=1cm, minimum height=0.5cm, align=center, draw=black, fill=purple!20,font=\scriptsize},
        output_box/.style={rectangle, rounded corners=4pt, minimum width=1cm, minimum height=0.5cm, align=center, draw=black, fill=yellow!20,font=\scriptsize},
        arrow/.style={-latex, thick},
        rounded corners=5pt  
    ]
    
    \coordinate (origin) at (0,0);
    \coordinate (origin_Q) at ([xshift=-0.5cm,yshift=0.0cm]origin);
    \coordinate (origin_K) at ([xshift=+0.0cm,yshift=0.0cm]origin);    
    \coordinate (origin_V) at ([xshift=+0.5cm,yshift=0.0cm]origin);
    \node (cir_encoder) [input_box, above left=of origin] {CIR\\Encoder};
    \node (rssi_encoder) [input_box, above right=of origin] {RSSI\\Encoder};
    \node (cir_input) [above=of cir_encoder,font=\scriptsize] {CIR $h_t$};
    \node (rssi_input) [above=of rssi_encoder,font=\scriptsize] {Fixed RSSI Map};
    \node (attention) [process_box, below=of origin] {Multi-Head Attention};
    \coordinate (attention_Q) at ([xshift=-0.5cm,yshift=0.0cm]attention.north);
    \node (attention_Q_label) [above left =0.075cm of attention_Q,font=\scriptsize] {Q};
    \coordinate (attention_K) at ([xshift=+0.0cm,yshift=0.0cm]attention.north);
    \node (attention_K_label) [above right =0.075cm of attention_K,font=\scriptsize] {K};    
    \coordinate (attention_V) at ([xshift=+0.5cm,yshift=0.0cm]attention.north);
    \node (attention_V_label) [above right =0.075cm of attention_V,font=\scriptsize] {V};
    \node (ssm) [core_box, below=0.5cm of attention.south, minimum width=3.8cm] {SSM Core\\$\mathbf{s}_t = \mathbf{A}\mathbf{s}_{t-1} + \mathbf{B}\mathbf{u}_{t}^{a} + \mathbf{w}_t$};
    \node (previous_state) [left=of ssm,font=\scriptsize] {$\mathbf{s}_{t-1}$};
    \node (next_state) [right=of ssm,font=\scriptsize] {$\mathbf{s}_{t}$};
    \coordinate (split) at ([xshift=0.0cm,yshift=-0.5cm]ssm.south);
    \node (decoder_3) [decoder_box, below=of split] {Decoder 3};
    \node (decoder_2) [decoder_box, left=0.3cm of decoder_3] {Decoder 2};
    \node (decoder_1) [decoder_box, left=0.3cm of decoder_2] {Decoder 1};
    \node (and_so_on) [right=of decoder_3] {...};
    \node (decoder_n) [decoder_box, right=of and_so_on] {Decoder $N$};
    \node (gating) [process_box, right=0.3cm of decoder_n,minimum width=0.3cm] {$f$};
    \coordinate (decoder_1_arrow_end) at ([xshift=0.0cm,yshift=-0.5cm]decoder_1.south);
    \coordinate (decoder_2_arrow_end) at ([xshift=0.0cm,yshift=-0.5cm]decoder_2.south);
    \coordinate (decoder_3_arrow_end) at ([xshift=0.0cm,yshift=-0.5cm]decoder_3.south);
    \coordinate (decoder_n_arrow_end) at ([xshift=0.0cm,yshift=-0.5cm]decoder_n.south);
    \node (mixture) [output_box, below=0.5cm of decoder_3.south, minimum width=7cm] {Mixture of Experts};
    \node (out) [below=of mixture,font=\scriptsize] {$\tilde{y}_t$};
    
    \draw [arrow] (cir_input) -- (cir_encoder);
    \draw [arrow] (rssi_input) -- (rssi_encoder);
    \draw [arrow] (cir_encoder) |- (origin_Q) -- (attention_Q);
    \draw [arrow] (rssi_encoder) |- (origin_K) -- (attention_K);
    \draw [arrow] (rssi_encoder) |- (origin_V) -- (attention_V);
    \draw [arrow] (attention) -- (ssm) node[midway, right, font=\tiny] {$\mathbf{u}_{t}^{a}$};
    \draw [arrow] (previous_state) -- (ssm);
    \draw [arrow] (ssm) -- (next_state);
    \draw [arrow] (ssm) -- (split) -| (decoder_1);
    \draw [arrow] (ssm) -- (split) -| (decoder_2);
    \draw [arrow] (ssm) -- (decoder_3);
    \draw [arrow] (ssm) -- (split) -| (decoder_n);
    \draw [arrow, dashed] (ssm) -- (split) -| (gating);
    \draw [arrow] (decoder_1) -- (decoder_1_arrow_end);
    \draw [arrow] (decoder_2) -- (decoder_2_arrow_end);
    \draw [arrow] (decoder_3) -- (decoder_3_arrow_end);
    \draw [arrow] (decoder_n) -- (decoder_n_arrow_end);
    \draw [arrow, dashed] (gating) |- (mixture.east);
    \draw [arrow] (mixture) -- (out);
    
    \end{tikzpicture}
    \caption{Architecture of the proposed SSM-based localization model. The model processes CIR and RSSI map inputs through separate encoders, applies attention, and uses an SSM core for temporal modeling before decoding using a mixture of expert. $\mathbf{A}$ and $\mathbf{B}$ are the SSM weight matrices, $\mathbf{w}_t$ is a regularization noise vector, and $f$ is a MLP used to compute the mixture weights.}
    \label{fig:ssmfinalmodel}
\end{figure}
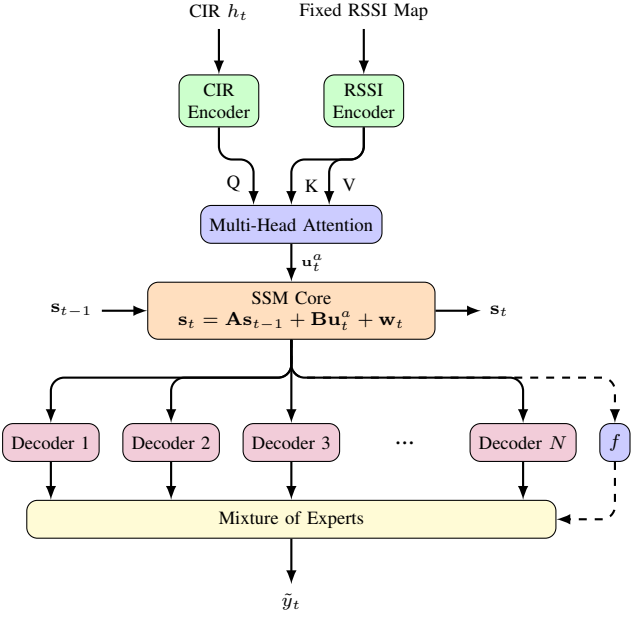

Our model is a structured state-space sequence model that predicts a spatial heatmap at each time step from radio measurements, optionally conditioned on a global prior (e.g., an RSSI coverage map). As illustrated in Figure~\ref{fig:ssmfinalmodel}, two encoder networks map each normalized CIR snapshot \(h_t\) and constant  RSSI prior to latent representations, then provided to an attention mechanism. A learned SSM backbone updates the latent state $s_t$ from $s_{t-1}$ to modulate the contribution of current evidence while preserving long-range memory. The updated state is decoded by a mixture-of-experts: multiple MLP decoders produce candidate heatmaps, and a mixer blend them into a final predicted distribution over the $H\times W$ grid.


\subsection{Loss and Metrics}

\subsubsection{Training loss}
At each time step \(t\), the model outputs a heatmap \(\tilde{y}_t\) normalized such that \(\sum_{i,j}\tilde{y}_t[i,j]=1\). The target \(y_t\) is the Gaussian-blurred probability map. We minimize the regularized Kullback–Leibler divergence:
\begin{equation}
    D_{\mathrm{KL}}\!\bigl(\mathbf{y}\| \tilde{\mathbf{y}}\bigr)+\left\|\varphi\right\|_2^2
    = \sum_{t=1}^T \sum_{i,j} y_t[i,j]\;\log\!\frac{y_t[i,j]}{\tilde{y}_t[i,j]}+\left\|\varphi\right\|_2^2.
    \label{eq:loss}
\end{equation}

\subsubsection{Evaluation metrics}\label{subsubsection:eval_metrics}

\paragraph{Cosine Similarity}
To assess spatial agreement between the predicted and target heatmaps at timestep $t$, we use the Cosine Similarity (CS):
\begin{equation}
\mathrm{CS}(\tilde{y}_t, y_t)
=
\frac{\left\langle \tilde{y}_t,\; y_t  \right\rangle}
{\left\|\tilde{y}_t \right\|_2 \;\left\|y_t\right\|_2}.
\end{equation}
CS is effective at capturing overall spatial pattern similarity. However, CS is highly sensitive to small spatial misalignments: even a one-grid-cell shift of a sharp peak can severely degrade the score, making it overly harsh for evaluating localization accuracy where small position errors may be practically acceptable.

\paragraph{Confidence Circles}
Let \(P_t=\{(i_k,j_k): b_t[i_k,j_k]=1\}\) be the set of active cells corresponding to binary ground‑truth heatmap $b_t$. For each active cell \(p_k=(i_k,j_k)\), we define a circle of radius \(\rho_k\) centered at \(p_k\). The radius $\rho_k$ is inversely proportional to RSSI at $p_k$. The union mask is \(M_t[i,j]=\mathds{1}\bigl(\exists k:\|(i,j)-p_k\|_2\le \rho_k\bigr)\): it equals 1 if there exists at least one circle covering cell $(i,j)$, otherwise 0. The score is the fraction of predicted mass lying inside the union of circles:
\begin{equation}
\mathrm{CC}(\tilde{y}_t,M_t)\;=\;\sum_{i,j}\tilde{y}_t[i,j]\; M_t[i,j].
\end{equation}

Complementary to the CS, this metric measures how predictions concentrate near ground-truth locations, with RSSI adaptive tolerance.

We emphasize that the task considered here is not direct detection or coordinate-level localization, but occupancy heatmap prediction. The predicted heatmap is intended as an intermediate probabilistic representation of spatial occupancy, which could serve as input to subsequent detection or localization stages in a complete sensing pipeline. Consequently, the proposed loss and evaluation metrics have been designed accordingly.

\section{Results and Analysis}\label{section:results}

\subsection{Experimental Setup}

We compare our SSM model with three baselines of comparable parameter counts ($\approx10^6$ to $10^7$ parameters): a Multi-Layer Perceptron (MLP), a Convolutional Neural Network (CNN), and a Vision Transformer (ViT). The MLP consists of fully-connected layers with ReLU activations. The CNN uses 1D convolutions with max-pooling. The ViT employs a two-stage architecture: spatial processing per timestep  followed by temporal attention across the sequence. 

Each model is trained and evaluated with k-fold cross-validation across two recording sessions, using five folds per session. Each fold is trained for 30 epochs with batch size 128, and the checkpoint with the best validation score is retained. Training samples are sequences of 10 time steps, and we consider scenarios with 1 to 3 moving persons in the room. The combinatorial space of configurations (16×16 positions, 1 to 3 occupants, continuous coordinates, random trajectories) vastly exceeds our total dataset size of 2000 sequences. Successful generalization therefore requires learning temporal dynamics and target-clutter separation rather than memorizing input-output pairs (overfitting).

\begin{figure*}[htbp]
    \centering
    \resizebox{1\textwidth}{!}{
        \begin{tikzpicture}
            \node[anchor=south west, inner sep=0] (image) at (0,0) {
                \includegraphics[width=1\linewidth, trim=0 0 0 60pt, clip]{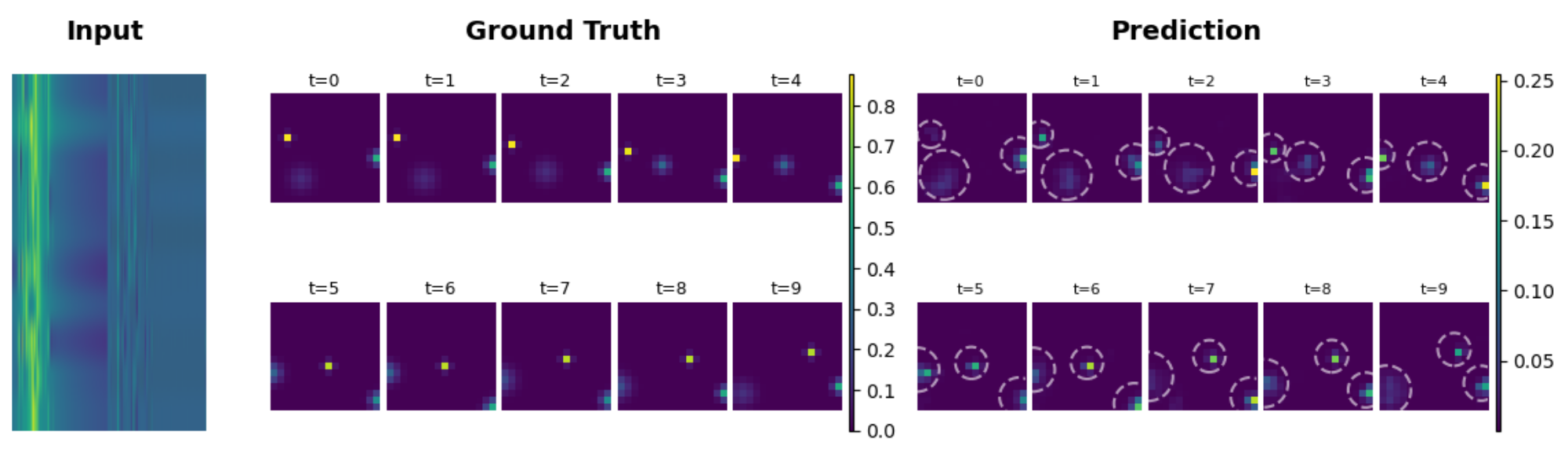}
            };
            
            \node[anchor=north] at (1.2, 4.9) {Inputs};
            \node[anchor=north] at (6.45, 4.9) {Ground Truth};
            \node[anchor=north] at (13.85, 4.9) {Predictions};
            \draw[decorate, decoration={brace, amplitude=5pt}, thick] (1.24, 0.2) -- (0.15, 0.2) node[midway, below=6pt, font=\footnotesize] {Amplitude};
            \draw[decorate, decoration={brace, amplitude=5pt}, thick] (2.37, 0.2) -- (1.26, 0.2) node[midway, below=6pt, font=\footnotesize] {Phase};
            \draw[latex-, thick] (0, 0.27) -- (0, 4.35) node[midway, left=6pt] {$t$};
            
        \end{tikzpicture}
    }
    \caption{Example predictions of our SSM model in the U‑shaped room. Target heatmaps are blurred as proposed in Section~\ref{subsubsection:label_blurring}. The circles overlaid on the predictions represent the confidence circles around the true targets (defined in Section~\ref{subsubsection:eval_metrics}). Predictions should concentrate within these circles.}
    \label{fig:viz_uroom_convergence}
\end{figure*}
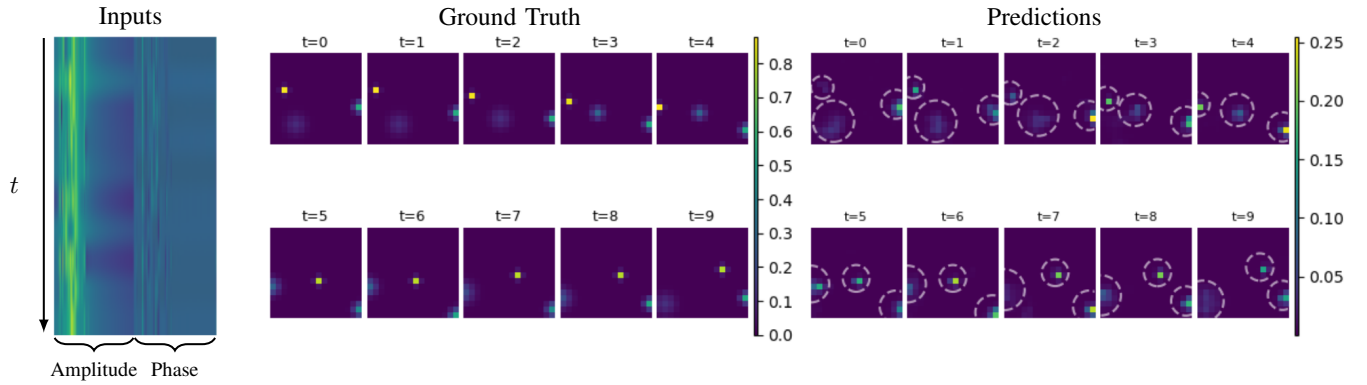

\subsection{Performance Summary}

Figure~\ref{fig:viz_uroom_convergence} illustrates an example in the U‑shaped room at convergence. Each CIR frame is represented as a two‑channel input (amplitude and phase in $\mathbb{R}^2$). The SSM processes a sequence of CIRs and outputs a probability heatmap per time step (right panel). The ground‑truth occupancy (middle panel) indicates three subjects present. As the sequence unfolds, the SSM’s heatmaps become progressively sharper, reflecting the integration of temporal evidence across time steps. This highlights the central role of sequence modeling in our approach.
\newcolumntype{C}{>{\centering\arraybackslash}X}
\newcolumntype{M}[1]{>{\centering\arraybackslash}m{#1}}
\renewcommand{\arraystretch}{1.4}

\begin{table}[htbp]
\centering
\caption{Performance Comparison}
\label{tab:performance_summary}
\begin{tabularx}{\columnwidth}{|>{\centering\arraybackslash}m{1.5cm}||C|C||C|C|}
\hline
\multicolumn{1}{|>{\centering\arraybackslash}m{1.5cm}||}{\textbf{Model}} & \multicolumn{2}{c||}{\textbf{Square Room}} & \multicolumn{2}{c|}{\textbf{U‑Room}} \\
\cline{2-5}
 & \textbf{CC (\%)} & \textbf{CS (\%)} & \textbf{CC (\%)} & \textbf{CS (\%)} \\
\hline
SSM & \textbf{51.6} & \textbf{94.3} & \textbf{48.3} & \textbf{95.3} \\
ViT & 49.6 & 82.2 & 43.4 & 74.1 \\
CNN & 28.2 & 60.2 & 20.9 & 53.9 \\
MLP & 27.3 & 39.4 & 25.3 & 40.4 \\
\hline
\end{tabularx}
\end{table}


Table~\ref{tab:performance_summary} summarizes the results. All scores are computed on the last heatmap. Across both environments, the SSM consistently achieves the best performance for both metrics, followed by the ViT, the CNN and the MLP. This gap suggests that explicit temporal modeling of CIR sequences adds information that purely spatial or shallow temporal models do not fully capture. On the U‑shaped room, scores generally decrease due to increased multipath and non‑LOS conditions, but the SSM maintains its relative advantage. Although larger studies are needed to confirm these findings across a broader range of environments and datasets, the present results identify the SSM and the ViT as particularly promising candidate architectures.

\section{Conclusion and Perspectives}\label{section:conclusion}

We presented a DT-in-the-loop methodology for scalable radio sensing and studied its feasibility through a PoC for DFL. Trained purely on synthetic CIR sequences generated via Sionna RT from 3D environment models, our SSM outperforms MLP, CNN, and ViT baselines across two geometries. While this study is simulation-only and restricted to two synthetic rooms, these results validate the core hypothesis: site-specific models trained on DT-generated data can learn to map CIR temporal sequences to probabilistic occupancy.



The main limitation is that the study remains simulation-only. The key next step is therefore real-world validation, to assess how DT fidelity (geometry precision, material accuracy, antenna modeling, ray tracing quality, hardware impairments) affects sim-to-real transfer. Beyond this, we plan to expand datasets to diverse geometries, occupancy scenarios, and environmental conditions; develop decision layers to convert probabilistic outputs into discrete detections, enabling standard detection metrics (false alarms, missed detections, etc.);  produce sensing-accuracy and coverage maps to optimize transmitter placement jointly for communication and sensing objectives; and assess whether models can generalize across environments or require per-site fine-tuning.


\bibliographystyle{IEEEtran}
\bibliography{biblio}

\vspace{12pt}

\end{document}